\pdfoutput=1
\documentclass[11pt]{article}
\usepackage[]{ACL2023}
\usepackage{times}
\usepackage{latexsym}
\usepackage[T1]{fontenc}
\usepackage[utf8]{inputenc}
\usepackage{microtype}
\usepackage{inconsolata}
\usepackage{graphicx}

\title{Towards Enhanced RAC Accessibility: Leveraging Datasets and LLMs}

\author{Bejarano Sepulveda Edison,\\ {\normalsize Universidad de Barcelona} \\ {\normalsize ejbejaranos@gmail.com} 
         \And Potes Hector Nicolai,\\ {\normalsize nicolai.potes00@gmail.com}
         \And Pineda Montoya Santiago,\\ {\normalsize Universidad Nacional de Colombia} \\ {\normalsize sapinedamo@unal.edu.co}
         \AND  Rodriguez Felipe Ivan,\\ {\normalsize Fund. Univ. Los Libertadores} \\ {\normalsize ifrodriguezb@libertadores.edu.co}
         \And  Orduy Jaime Enrique,\\ {\normalsize Fund. Univ. Los Libertadores} \\ {\normalsize jaime.orduy@libertadores.edu.co}
         \AND  Rosales Cabezas Alec,\\ {\normalsize Fund. Univ. Los Libertadores} \\ {\normalsize amrosalesc@libertadores.edu.co}
         \And  Traslaviña Navarrete Danny\\ {\normalsize Fund. Univ. Los Libertadores} \\ {\normalsize dstraslavinan@libertadores.edu.co}
         \And  Madrid Farfan Sergio,\\ {\normalsize Fund. Univ. Los Libertadores} \\ {\normalsize snmadridf@libertadores.edu.co}
         }

\begin{document}
\maketitle
\begin{abstract}

This paper explores the potential of large language models (LLMs) to make the Aeronautical Regulations of Colombia (RAC) more accessible. Given the complexity and extensive technicality of the RAC, this study introduces a novel approach to simplifying these regulations for broader understanding. By developing the first-ever RAC database, which contains 24,478 expertly labeled question-and-answer pairs, and fine-tuning LLMs specifically for RAC applications, the paper outlines the methodology for dataset assembly, expert-led annotation, and model training. Utilizing the Gemma1.1 2b model along with advanced techniques like Unsloth for efficient VRAM usage and flash attention mechanisms, the research aims to expedite training processes. This initiative establishes a foundation to enhance the comprehensibility and accessibility of RAC, potentially benefiting novices and reducing dependence on expert consultations for navigating the aviation industry's regulatory landscape.

You can visit the \href{https://huggingface.co/somosnlp/gemma-1.1-2b-it_ColombiaRAC_FullyCurated_format_chatML_V1}{dataset} and the \href{https://huggingface.co/datasets/somosnlp/ColombiaRAC_FullyCurated}{model} here.


\end{abstract}

\section{Introduction}
\paragraph{}

The Colombian aviation industry operates under the Aeronautical Regulations of Colombia (RAC) \cite{rac2023}, a comprehensive legal framework that comprises approximately 50 detailed regulations and manuals. It is worth noting that the RAC is currently undergoing harmonization with the Latin American Aeronautical Regulations (LAR). The technical complexity and voluminous nature of these documents pose significant challenges to accessibility. However, the advent of LLMs promises to revolutionize this scenario by simplifying complex texts, making regulatory information more understandable and accessible to a broader audience. As highlighted by Yang et al. \cite{yang2023harnessing}, LLMs such as ChatGPT have demonstrated significant potential in various practical applications, suggesting their utility in interpreting and simplifying legal and regulatory texts. By translating legal terminology into plain language, LLMs play a crucial role in demystifying aviation regulations, thereby improving understanding and compliance within the industry.

\section{Objectives}



The overarching goal is to utilize LLMs to make the Aeronautical Regulations of Colombia (RAC) more accessible and understandable for both aviation professionals and the general public. The approach is multifaceted: First, we aim to develop a comprehensive dataset from the RAC's initial five documents, laying the groundwork for LLM training. Second, this project seeks collaboration with industry and academic experts to annotate and refine this dataset, ensuring its relevance and accuracy. Third, we will train LLMs with the curated dataset, and lastly, evaluate their performance in simplifying the RAC's content based on feedback from aeronautical experts, thereby enhancing regulatory compliance and understanding.

\section{Background}
Innovative tech solutions are crucial for navigating the complexities of the Aeronautical Regulations of Colombia (RAC). Starting with decision trees to enhance access to RAC, the aviation sector's push for safety and efficiency demands more sophisticated approaches, given its high reliability standards \cite{Rodriguez2021, IEEEInnovation}. Early AI uses in aviation, such as expert systems for decision support, highlight AI's impact on safety and operational efficiency \cite{EASARoadmap}. The emergence of LLMs, including Seq2Seq and Transformer architectures, marks a significant advancement, offering detailed conversational AI support \cite{MDPIArticle, CMSConferences, IEEESAFARI}. A UAEAC (Unidad Administrativa Especial de Aeronáutica Civil) project demonstrated AI's potential for real-time RAC inquiries, showcasing AI's role in enhancing regulatory consultations \cite{UNADStudy}. This move towards LLMs for regulatory compliance and decision support illustrates a shift to intelligent solutions, with LLMs providing the flexibility and depth to address the regulatory domain's complexities effectively.

\section{Methodology}
This research systematically applies LLMs to enhance the accessibility of the Aeronautical Regulations of Colombia (RAC). Our methodology includes dataset generation, expert-driven labeling, and iterative LLM fine-tuning, ensuring a comprehensive, data-driven approach.

\subsection{Dataset Generation from RAC Documents}

\begin{figure*}[ht!]
    \centering
    \includegraphics[width=14cm, height=6cm, keepaspectratio=false]{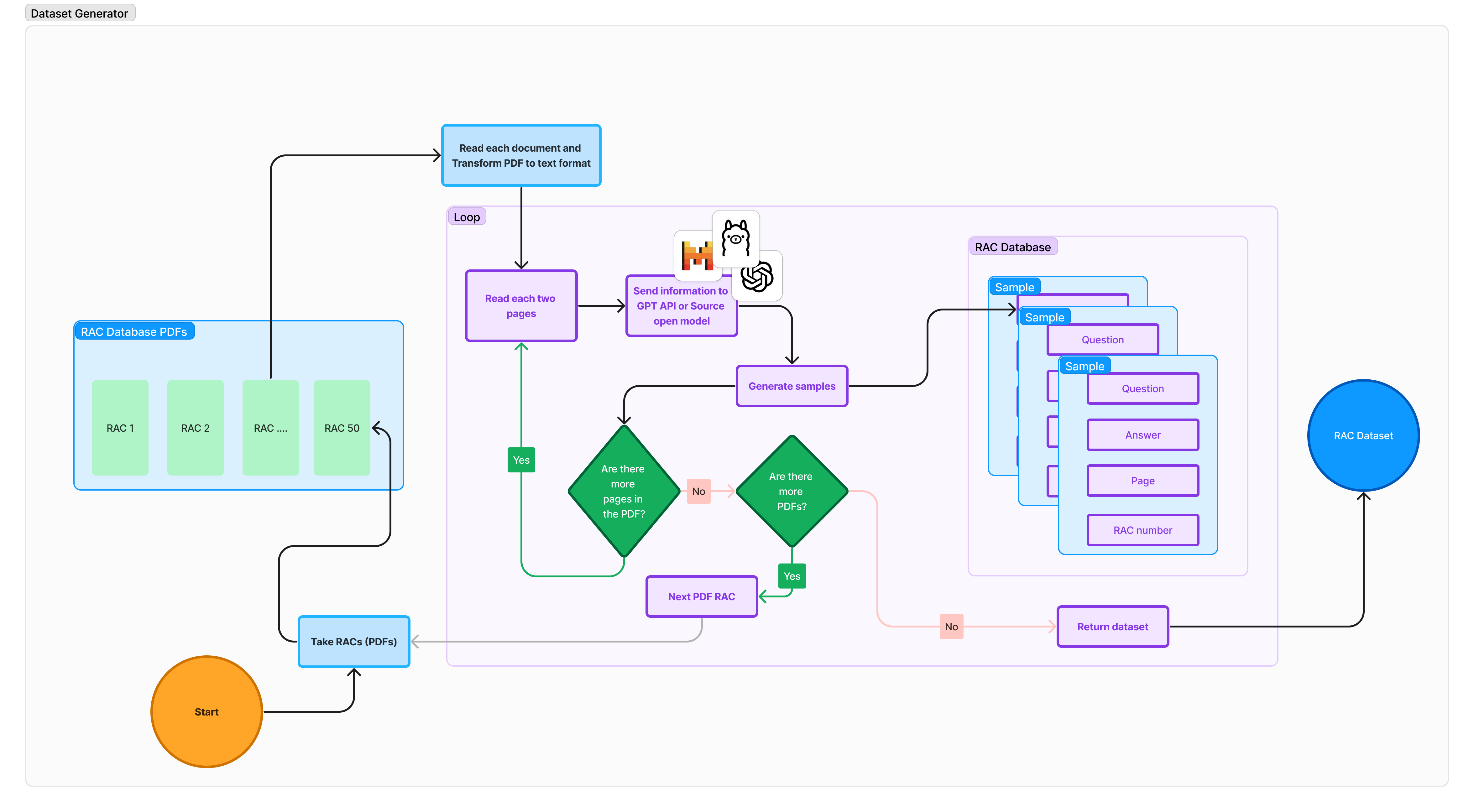}
    \caption{Process for data extraction from RAC PDFs using GPT API.}
    \label{fig:dataset_diagram}
\end{figure*}

The dataset was crafted from the RAC using an automated process, as depicted in Figure~\ref{fig:dataset_diagram}. This process began with converting PDFs to text, followed by processing through a GPT API or a similar model. The system handled two pages at a time and iteratively compiled the extracted data. The resulting dataset, comprising questions, answers, and relevant RAC references, was thus assembled for further analysis.

\subsection{Labeling Process for the RAC Dataset}





\begin{figure}[ht!]
    \centering
    \includegraphics[width=0.9\linewidth, height=8cm]{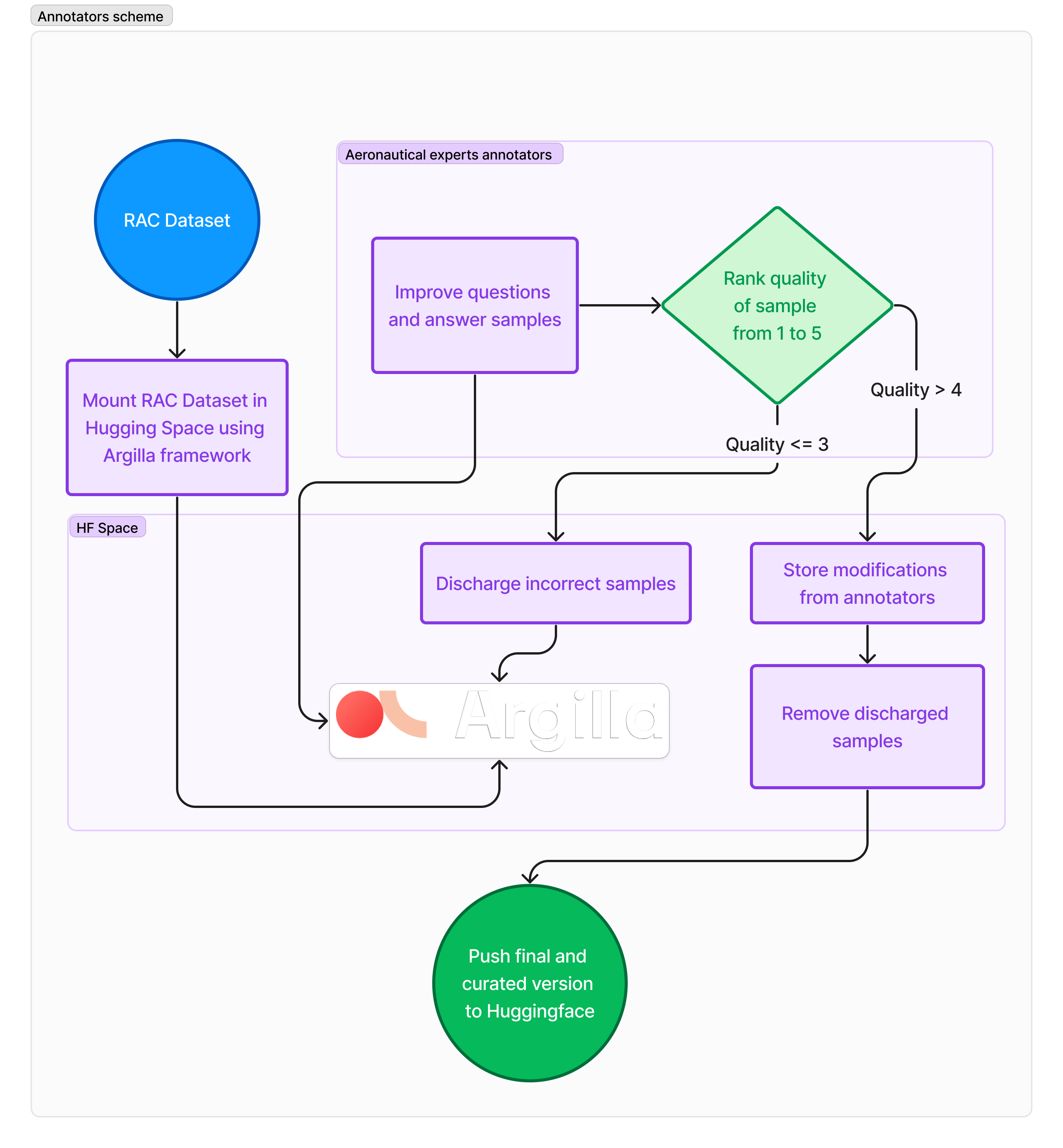}
    \caption{Flow diagram for system annotator.}
    \label{fig:annotator_diagram}
\end{figure}

The RAC Dataset underwent refinement using the Argilla framework within the Hugging Face environment, as illustrated in Figure \ref{fig:annotator_diagram}. This tool facilitated structured annotation tasks, leveraging the expertise of aeronautical engineering specialists from Fundación Universitaria Los Libertadores. They assessed each sample for quality, retaining those ranked above 3 and removing lower-ranked ones, thus ensuring the dataset's integrity.

 The concluding phase of the process involved removing the discarded samples. Consequently, a high-quality dataset was consolidated and made available through Hugging Face.

\subsection{Fine tunning models}

In this study, we fine-tuned the pre-trained GEMMA model for a specific NLP task by employing a Parameter-Efficient Fine-Tuning (PEFT) strategy integrated with Low-Rank Adaptation (LoRA). Initially, the somosnlp/ColombiaRACFullyCurated dataset was tokenized and divided into training and testing subsets. 

\begin{table}[ht]
\small
\centering
\begin{tabular}{lll}
\hline
\textbf{Version} & \textbf{Data Quality} & \textbf{Model Type} \\
\hline
V8 & Clean & gemma-1.1-2b-it \\
V7 & Clean & gemma-1.1-2b-it \\
V6 & 50\% Clean & gemma-2b-it \\
V5 & 50\% Clean & gemma-2b-it \\
V4 & 50\% Clean & gemma-2b-it \\
V3 & Raw & gemma-2b-it \\
V2 & Raw & gemma-2b-it \\
V1 & Raw & gemma-2b-it \\
\hline
\end{tabular}
\caption{Model Versions, Quality, and Types}
\label{tab:methodology1}
\end{table}

The model was then configured to efficiently adapt to the task by selectively modifying a minimal subset of its parameters, utilizing PEFT and LoRA techniques. Key hyperparameters for the training included a learning rate of 5e-5 and a batch size of 3, optimized using the AdamW optimizer with a weight decay of 0.001.

\begin{table}[ht]
\small
\centering
\begin{tabular}{lll}
\hline
\textbf{Version} & \textbf{Techniques} & \textbf{GPU Type} \\
\hline
V8 & flashattn+perft+qlora & RTX 3090 \\
V7 & unsloth+peft+qlora & L4-24GB \\
V6 & unsloth+peft+qlora & L4-24GB \\
V5 & unsloth+peft+qlora & A100-40GB \\
V4 & peft+qlora & A100-40GB \\
V3 & peft+qlora & A100-40GB \\
V2 & peft+qlora & A100-40GB \\
V1 & peft+qlora & A100-40GB \\
\hline
\end{tabular}
\caption{Summary of Techniques and GPU Type}
\label{tab:techniques_gpu}
\end{table}

The training, managed by the SFTTrainer, emphasized gradient accumulation and learning rate scheduling to enhance model performance. After training, the fine-tuned model was combined with LoRA weights for deployment. This process demonstrates an efficient approach to customizing large-scale language models for specific tasks, leveraging advanced techniques to balance computational efficiency and model effectiveness. PyTorch and Hugging Face transformers were utilized for implementation due to their robustness and support for complex NLP tasks.

\section{Results}

\subsection{Quantitative results}
The quantitative results (see Table~\ref{tab:results}) highlight the effectiveness of the efforts. Earlier versions (V5 to V3) had low loss but lacked answer quality, while subsequent iterations improved response accuracy, training efficiency, and environmental sustainability. V8 was the most optimized model, enhancing RAC interpretability with advanced LLMs, lower training loss, and reduced environmental impact.

\begin{table}[ht]
\small
\centering
\begin{tabular}{llll}
\hline
 & \textbf{Time(s)} & \textbf{Loss} & \textbf{FLOPs} \\
\hline
V8 & 12,607 & 0.194 & 3.94E+14 \\
V7 & 33,262 & 0.194 & 2.51E+15 \\
V6 & 1,977 & 0.243 & 5.01E+15 \\
V5 & 1,779 & 0.076 & 5.18E+15 \\
V4 & 1,833 & 0.092 & 5.06E+15 \\
V3 & 3,987 & 0.071 & 4.99E+15 \\
V2 & 4,239 & 1.241 & 4.95E+15 \\
V1 & 50,973 & 0.6 & 2.29E+15 \\
\hline
\end{tabular}
\caption{Summary of Results: Runtime, Loss, FLOPs}
\label{tab:results}
\end{table}

\subsection{Qualitative results}

Table \ref{tab:mean_median_test_count} shows the model's strong performance with average scores of 7 from 276 tests. However, RAC 3's low scores (mean 3.464, median 1) indicate areas needing improvement, while high ratings in RACs 1 and 5 suggest strengths. These results confirm the model's potential for accuracy and generalization, though RAC 3 requires adjustments.

\begin{table}[ht]
\small
\centering
\begin{tabular}{llll}
\hline
\textbf{RAC} & \textbf{Mean} & \textbf{Median} & \textbf{Number of Tests} \\
\hline
1 & 6.12 & 7 & 100 \\
2 & 5.8 & 6.5 & 20 \\
3 & 3.464 & 1 & 56 \\
5 & 6.34 & 7.5 & 100 \\
General & 5.6 & 7 & 276 \\
\hline
\end{tabular}
\caption{Expert Evaluations by RAC: Mean, Median, and Test Count}
\label{tab:mean_median_test_count}
\end{table}

\section{Conclusions}
The overarching goal was to enhance accessibility to the RAC through the utilization of LLMs. This objective was pursued through a multi-faceted approach: first, by developing a comprehensive dataset from the initial RAC documents; second, by collaborating with industry and academic experts to refine the dataset; third, by training LLMs with this curated dataset; and finally, by evaluating their performance. This initiative aims to improve regulatory compliance and understanding among aviation professionals and the general public.

\clearpage


\bibliography{anthology,custom}
\bibliographystyle{acl_natbib}




\end{document}